\documentclass[letterpaper, 10 pt, conference]{ieeeconf}  

\IEEEoverridecommandlockouts         
\usepackage{amsmath}
\usepackage{amssymb}
\usepackage{soul}
\usepackage{algorithm}
\usepackage{algpseudocode}

\usepackage{hyperref}

\usepackage{epsfig}
\usepackage{graphicx}
\usepackage{subcaption}

\usepackage{booktabs}
\usepackage{pifont}
\usepackage[table,x11names]{xcolor}

\usepackage{multicol}
\usepackage{multirow}

\newcommand{\mbf}[1]{\mathbf{#1}}

\definecolor{LightCyan}{rgb}{0.88,1,1}

\usepackage{graphics} 
\usepackage{amsmath} 
\usepackage{amssymb}  
\usepackage{graphicx} 
\usepackage{caption}
\usepackage{comment}
\usepackage{subcaption}

\title{\LARGE \bf
Robust visual sim-to-real transfer for robotic manipulation
}

\author{Ricardo Garcia$^1$, Robin Strudel$^1$,  Shizhe Chen$^1$, Etienne Arlaud$^1$, Ivan Laptev$^1$ and Cordelia Schmid$^1$
\thanks{$^1$Inria, \'Ecole normale sup\'erieure, CNRS, PSL Research University, 75005, Paris, France. {\tt\small \{firstname.lastname\}@inria.fr}}
}

\begin{document}

\bstctlcite{IEEEexample:BSTcontrol}

\twocolumn[{%
\renewcommand\twocolumn[1][]{#1}%
\maketitle
\pagestyle{plain}

\vspace{-2.1em}
\begin{center}
\includegraphics[width=0.95\linewidth]{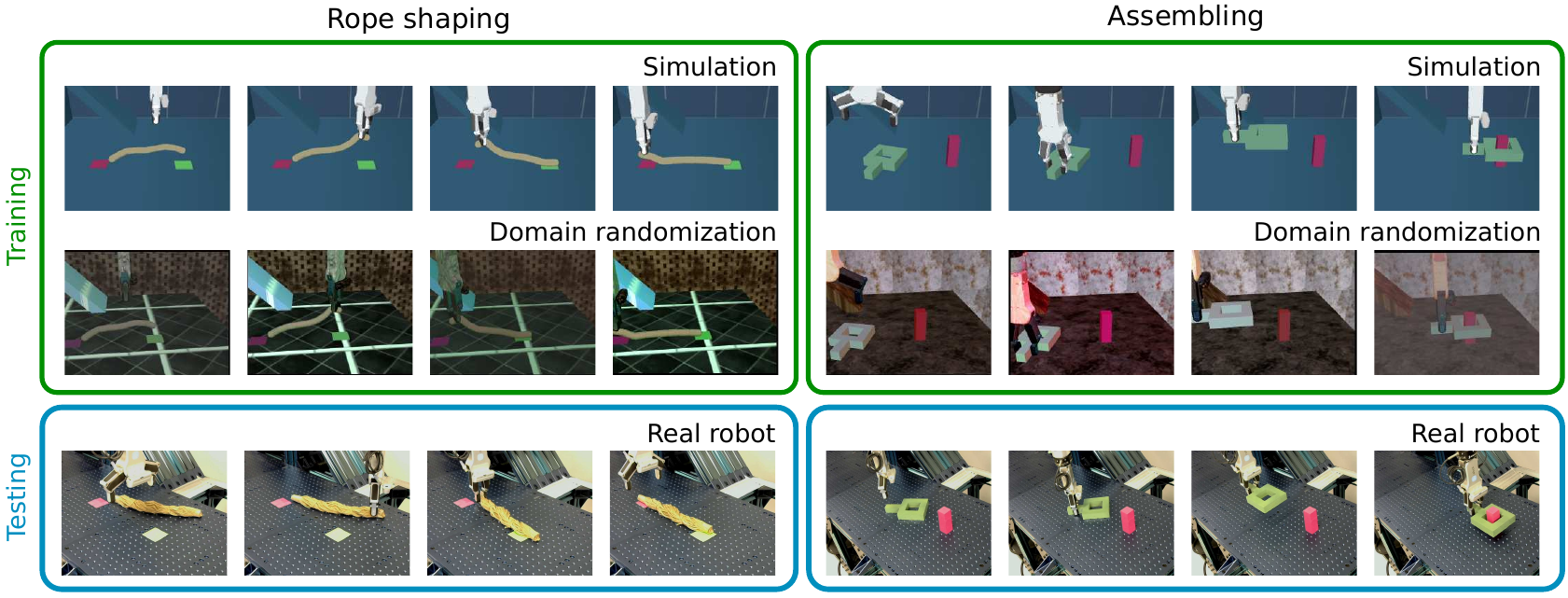}
\captionof{figure}{\textbf{An overview of our approach.} We learn visuomotor manipulation policies in simulation (row 1) with domain randomization by sampling high-quality textures, lighting, object colors and camera parameters (row 2). We analyze different sampling options and demonstrate that simulator-trained policies can be directly deployed on a real robot for diverse and challenging manipulation tasks (row 3), such as rope-shaping (left) and assembling (right). 
}
\label{fig:teaser}
\end{center}
}]

\begin{abstract}
Learning visuomotor policies in simulation is much safer and cheaper than in the real world. However, due to discrepancies between the simulated and real data, simulator-trained policies often fail when transferred to real robots. 
One common approach to bridge the visual sim-to-real domain gap is domain randomization (DR). 
While previous work mainly evaluates DR for disembodied tasks, such as pose estimation and object detection, 
here we systematically explore visual domain randomization methods and benchmark them on a rich set of challenging robotic manipulation tasks.
In particular, we propose an off-line proxy task of cube localization to select DR parameters for texture randomization, lighting randomization, variations of object colors and camera parameters.   
Notably, we demonstrate that DR parameters have similar impact on our off-line proxy task and on-line policies.
We, hence, use off-line optimized DR parameters to train visuomotor policies in simulation and directly apply such policies to a real robot. 
Our approach achieves 93\% success rate on average when tested on a diverse set of challenging manipulation tasks.
Moreover, we evaluate the robustness of policies to visual variations in real scenes and show that our simulator-trained policies outperform policies learned using real but limited data.
Code, simulation environment, real robot datasets and trained models are available at \url{https://www.di.ens.fr/willow/research/robust_s2r/}.

\footnotetext[1]{Inria, \'Ecole normale sup\'erieure, CNRS, PSL Research University, 75005, Paris, France.}

\vspace{-.1cm}
\end{abstract}

\section{Introduction}
\vspace{-.1cm}

Data-driven policy learning allows to acquire reactive robotic skills and has recently shown
impressive results both in simulation and in the real world for tasks such as dexterous manipulation \cite{akkaya19, allshire21, handa22, chen22}, robotic arm manipulation~\cite{levine16, lee21, florence21}, quadruped locomotion \cite{miki22,frey22,agarwal22,aractingi22} and navigation \cite{chaplot2020object,chen2021history,chen2022think,khandelwal2022simple}. 
Despite much progress in recent years, deploying simulator-trained policies in the real world remains challenging due to the visual and physical sim-to-real gap \cite{akkaya19,handa22,miki22,bousmalis18,agarwal22}. 
As result, current methods often rely on real-robot training data~\cite{lee21,florence21,bousmalis18,florence20,zhang18} and are hard to scale due to the prohibitive cost of large-scale real data collection.

To reduce the visual sim-to-real gap, existing methods typically apply domain adaptation (DA)~\cite{bousmalis18, rao20, ho21} or domain randomization (DR)~\cite{handa22,lee21,tobin17,strudel20} techniques.
While DA attempts to create more realistic synthetic images, DR merely randomizes scene parameters such as textures and lighting, and is more common due to its simplicity and good performance in practice.
Nevertheless, most existing DR methods study sim-to-real transfer for disembodied tasks such as pose estimation~\cite{tobin17, alghonaim21} and object detection~\cite{valtchev20} as they are easier to evaluate than real robot tasks.
Hence, it remains unclear
(i)~whether DR generalizes well to diverse and challenging robotic tasks,
(ii)~how to select optimal DR parameters for learning robotic policies, and 
(iii) how robotic policies trained with DR in simulation compare to policies trained on real data when tested in real scenes with varying visual appearance.

In this work we address the above questions and systematically study visual domain randomization on a suite of seven challenging robotic manipulation tasks, see Figure~\ref{fig:overview}. 
Our initial experiments show that policies trained in simulation without adaptation fail and achieve zero accuracy when transferred to a real robot.
To improve sim-to-real transfer, we propose a cube localization proxy task and use it for off-line optimization of DR parameters such as textures, lighting, object colors and camera parameters.
We then use DR with the obtained set of parameters and train policies for seven robotic manipulation tasks, namely stacking, box-retrieving, assembling, pushing, pushing-to-pick, sweeping and rope-shaping, see examples in Figure~\ref{fig:teaser}.

As one of our main contributions, we demonstrate that improvements on our proxy task consistently transfer to all considered manipulation policies under the same DR parametrization.
Our policies trained on simulation-only data using off-line optimized DR parameters achieve 93\% average success rate on seven real-world manipulation tasks.
Moreover, we show that our simulation-trained policies demonstrate robustness to appearance variations of real scenes and outperform policies trained on real but limited data.

In summary, our contributions are three-fold: 
(i)~We propose a proxy task and demonstrate that DR parameters optimized on this task generalize to a rich set of manipulation policies.
(ii)~We present a diverse set of seven manipulation tasks and use them to benchmark sim-to-real policy transfer.
(iii)~We ablate and show high success rate for policies trained only in simulation across a diverse set of manipulation tasks on a real robot. 
We also demonstrate our approach to outperform policies trained on real but limited data.

\section{Related Work}

The discrepancies between simulation and the real world mainly come from the visual appearance of the scene \cite{denninger20,sadeghi18,qin22} and the 
physical dynamics \cite{xue18,valassakis20,mehta20,tsai21}. Here, we focus on reducing the visual sim-to-real gap and review the two dominant approaches, namely domain adaptation and domain randomization.

\noindent\textbf{Domain adaptation} methods aim to map images rendered in simulation to realistic images, for example using Generative adversarial networks (GANs) \cite{bousmalis18, rao20, ho21}.
Alternatively, such methods attempt to map simulated and real data to a common domain either in the image space~\cite{james19} or feature space~\cite{ganin16}.
Other methods \cite{hansen21} perform adaptation by finetuning the model during deployment using self-supervision.
While GANs are expressive models, they often generate artifacts~\cite{bousmalis18} such as spurious objects and scene structures. 
To preserve the 3D structure of scenes in generated images, domain adaptation methods often resort to manually-defined constraints such as segmentation masks~\cite{bousmalis18,ho21}.

\noindent\textbf{Domain randomization} methods, also known as domain augmentation, focus on adding variation to the simulated data to enforce robustness of the learned model to image perturbations. 
Tobin \textit{et al.} \cite{tobin17} change the visual appearance of the simulation by randomizing the material textures, objects pose, shape and color, background, lighting properties and camera parameters. 
While this method can successfully transfer robotic policies trained in simulation to real robots~\cite{james17,andrychowicz20,matas18,sadeghi18}, it requires a significant amount of manual tuning and expensive iterations of simulation-based policy training followed by real-robot validation.  
For this reason, systematic and efficient domain randomization is still an open question~\cite{hofer20}. In this work, we address this question by examining how domain randomization parameters should be chosen to make them usable across different tasks and scene conditions.
Alghonaim and Johns~\cite{alghonaim21} study a set of design choices for visual sim-to-real. However, the evaluation is limited to one off-line task, object pose estimation. Our work goes beyond this task studying design choices for a  varied set of on-line robotic control tasks. 
Pashevich et al. \cite{pashevich19} exploit object localization as a proxy task to optimize 2D image augmentations for depth images with Monte-Carlo tree search (MCTS). 
In contrast, we address sim-to-real transfer for RGB images where our experimental results in Table~\ref{table:dr_ablation} confirm that 3D domain randomization is by far more effective than 2D image augmentations. The cost of 3D rendering prohibits MCTS optimization. We empirically show that a simple greedy search strategy on a proxy task enables successful sim-to-real transfer of visuomotor policies for diverse and challenging manipulation tasks.

\section{Approach}
Our goal is to train robust visuomotor policies in simulation and to deploy them  to manipulation tasks on a real robot.
We first describe our visual policies in Section~\ref{sec:bc} and introduce domain randomization for visual sim-to-real transfer in Section~\ref{sec:dr}. We then present our approach for selecting parameters of domain randomization using a proxy task in Section~\ref{sec:proxy_task}.

\begin{figure}[t]
  \centering
  \vspace{0.2cm}\includegraphics[width=\linewidth]{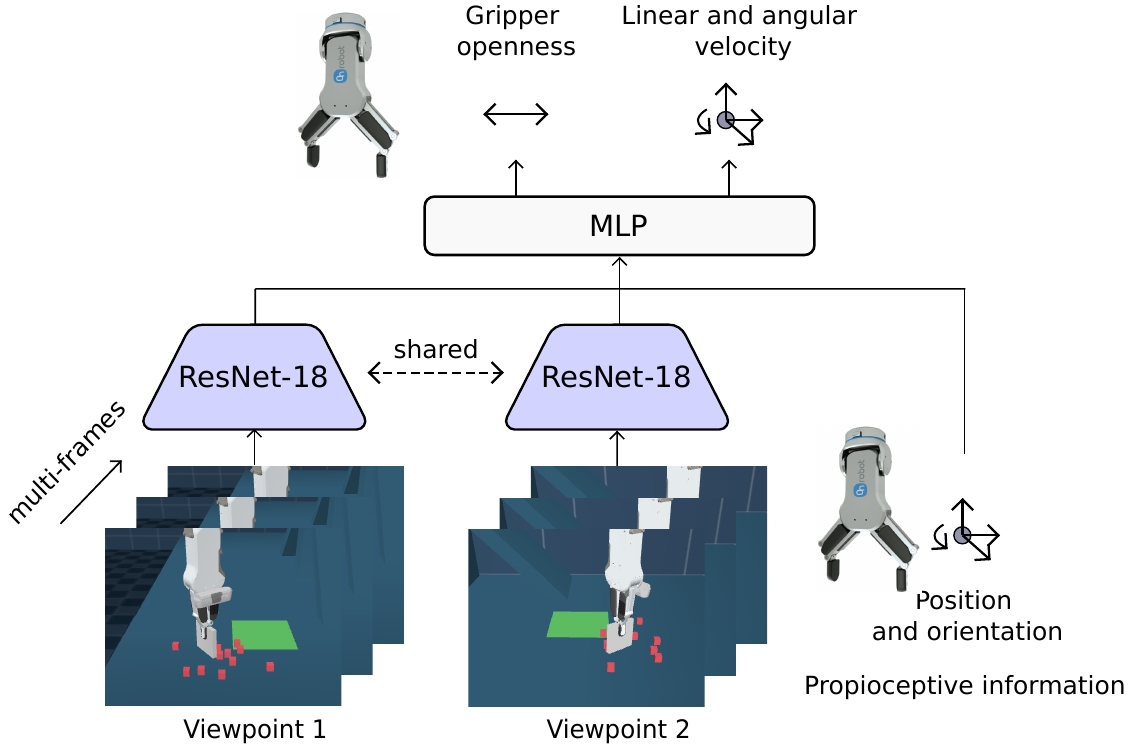}
  \caption{\textbf{Overall architecture of our policy network.} The network predicts the next action given a history of multiple frames from two viewpoints and the proprioceptive information of the gripper. }
  \label{fig:model_arch}
    \vspace{-.7cm}
\end{figure}

\vspace{-.1cm}

\subsection{Visuomotor policies for robotic manipulation}
\label{sec:bc}

\noindent\textbf{Problem formulation.}
We learn a policy $\pi_{\theta}(\mbf{a}_{t}\,|\,\mbf{o}_{t})$ that maps environment observation $\mbf{o}_{t}$ to a robot action $\mbf{a}_{t}$ at  timestep $t$, see Figure~\ref{fig:model_arch}.
The action $\mbf{a}_{t}=(\mbf{v}_{t}, \boldsymbol{\omega}_{t}, g_{t})$ is composed of the linear velocity $\mbf{v}_{t} \in \mathbb{R}^{3}$, the angular velocity $\boldsymbol{\omega}_{t} \in  \mathbb{R}^{3}$ of the robot end-effector and the gripper openness state $g_{t} \in \{0, 1\}$. 
The policy is executed in a closed-loop setting to perform a manipulation task. 

\noindent\textbf{Model architecture.}
Our model takes input from two RGB cameras as frames $\mbf{I}_{t} = \{\mbf{I}^1_{t}, \mbf{I}^2_t \}$ at each timestep $t$ where $\mbf{I}^{*}_{t} \in \mathbb{R}^{H \times W \times 3}$.
The cameras are placed on a wide baseline with 90 degrees relative angle to alleviate depth ambiguities and occlusions.  
The observation $o_t$ includes the last three RGB frames $(\mbf{I}_{t-2}, \mbf{I}_{t-1}, \mbf{I}_t)$ and the last three gripper proprioceptive values $\mbf{P}_{t}=[\text{pos}_t, \sin(\phi_t), \cos(\phi_t)]$, where $\text{pos}_t$ is the gripper position with respect to the robot base, 
and $\phi_t$ is the rotation angle of the gripper around the end-effector axis. 
We stack frames from each viewpoint in the channel dimension and use \mbox{ResNet-18} network~\cite{he16} to generate a feature vector of dimension $512$ for each viewpoint. 
We then concatenate feature vectors from the two viewpoints together with the proprioceptive information $\mbf{P}_{t}$.
The action $\mbf{a}_{t}$ is finally predicted by a multi-layer perceptron (MLP) composed of two layers with $512$ hidden units and a ReLU activation function each.

\noindent\textbf{Training with behavior cloning.}
Given a dataset of observation-action pairs $\{(\mbf{o}_{t}, \mbf{a}_{t})\}_{t}$ obtained from expert trajectories, 
we randomly initialize our policy network
and train it using a combination of the mean-squared error (MSE) loss for velocity control $\mbf{v}_{t}$, $\boldsymbol{\omega}_{t}$, and the binary cross entropy (BCE) loss for gripper state probability $g_{t}$ defined as:
\begin{equation}
  L = \lambda L_{MSE}((\hat{\mbf{v}}_{t}, \hat{\boldsymbol{\omega}}_{t}), (\mbf{v}_{t}, \boldsymbol{\omega}_{t})) + (1-\lambda)L_{BCE}(\hat{g}_{t}, g_{t})
  \label{eqn:loss}
\end{equation}
where $\lambda$ is a hyper-parameter to balance the MSE and BCE terms, $(\mbf{v}_{t}, \boldsymbol{\omega}_{t}, g_{t})$ is the expert action and $\pi_{\theta}(o_{t}) = (\hat{\mbf{v}}_{t}, \hat{\boldsymbol{\omega}}_{t}, \hat{g}_{t})$ is the predicted action for observation $o_t$.
\vspace{-.1cm}

\subsection{Domain Randomization}
\label{sec:dr}
In order to improve sim-to-real transfer, we augment synthetic images with domain randomization (DR) for policy learning.
We investigate different visual DR components, including textures, lightning conditions, object colors and camera parameters as described below.

\begin{figure}[t]
 \centering
 \vspace{0.2cm}\includegraphics[width=0.95\linewidth]{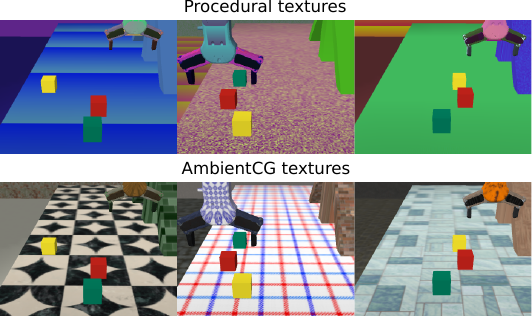}
 \caption{\textbf{Two types of textures in texture randomization.}}
  \label{fig:textures_comp}
    \vspace{-.5cm}
\end{figure}

\noindent\textbf{Texture randomization.} 
To obtain robustness to scene appearance, we randomize the textures of the robot, table, wall and floor. 
As the ability to distinguish object colors matters in our tasks, we do not randomize the object textures. 
We compare two types of textures as illustrated in Figure~\ref{fig:textures_comp}.
The first type is procedural textures \cite{tobin17} (Figure~\ref{fig:textures_comp} top), which have random colors and follow one of the four patterns - checkers, gradient, noise and a plain color. 
The second type is a set of $1,203$ high quality and realistic textures from ambientCG assets \cite{ambientCG} (Figure~\ref{fig:textures_comp} bottom). 

\noindent\textbf{Lighting randomization.} To achieve robustness to lighting conditions, we sample the light position uniformly in a portion of a sphere for rendering images. 
We define the sphere in spherical coordinates with a distance of the light source to the workspace center in the range $[1.0, 3.0]$ meter, azimuthal angle and polar angle in the range $[0, \pi/2]$ and $[\pi/10, 4\pi/10]$ radians. This range of parameters allows us to sample realistic light source positions around the table, e.g., we avoid sampling lights inside the table or under it.
Besides light positions, we randomize the light properties with diffuse, specular and ambient coefficients around a nominal value set to $0.3$ by adding an offset sampled from the range $[-\psi_{l}, \psi_{l}]$ where $\psi_{l}$ is a parameter to be optimized.

\noindent\textbf{Variation of object colors.} 
We treat object colors in a special way since object manipulation in some of our tasks depends on the object color.
We therefore randomize object colors by sampling the Hue, Saturation and Value (Brightness) (HSV) around their nominal value. 
We sample an offset from range $[-\phi_o, \phi_o]$ and add it to the object color expressed using HSV coordinates in $[0, 1]$. The range $\phi_o$ is selected such that it is sufficiently large to cover possible color discrepancies between real and simulated scenes, and not too large to avoid confusion between different objects with initially different colors.

\noindent\textbf{Variation of camera parameters.} Camera calibration is often imprecise especially in terms of extrinsic parameters. 
To improve robustness to slight viewpoint changes, we randomize camera positions in simulation by sampling camera angles, locations and the field of view (FOV) around default values.

\begin{figure*}[t]
 \centering
 \vspace{.2cm}
\includegraphics[width=\linewidth]{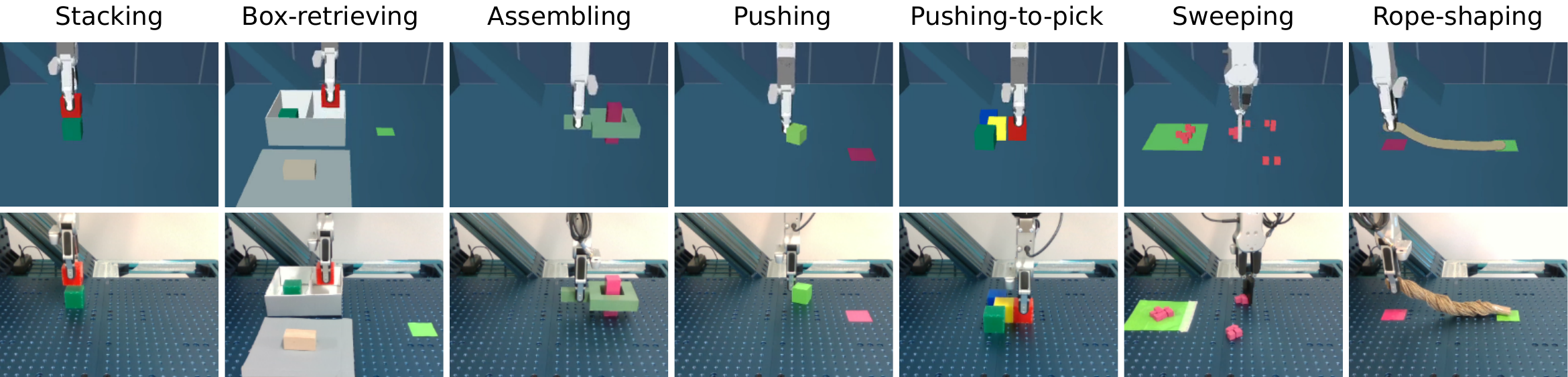}
 \caption{\textbf{The seven manipulation tasks considered in our experiments.}
 Images from the simulated environment (top) and our real robot environment (bottom) demonstrate the  gap between the two visual domains.   
 Our tasks exhibit a variety of challenges: 
 pick-and-place requires precise localization of objects and control of the end-effector; pushing is a contact-rich task that requires closed-loop reaction to adapt to object motion; box-retrieving is long-horizon; and rope-shaping involves manipulation of a deformable object.}
  \label{fig:overview}
  \vspace{-.6cm}
\end{figure*}
\vspace{-.2cm}
\subsection{Localization as a proxy task}
\vspace{-.1cm}
\label{sec:proxy_task}

DR parameters could be selected for each task by optimizing the accuracy of the learned policies for a real robot scenario. 
Such an approach, however, is highly time-consuming as it requires policy retraining and real-world policy evaluation for each set of DR parameters.
In this work we demonstrate that DR parameters can be efficiently chosen off-line using real images recorded for a proxy task. 
For this purpose we propose the following task of cube localization.

Given RGB images of a scene with three cubes of different colors, our proxy task aims to predict 3D locations of each cube relative to the gripper, see Figures~\ref{fig:textures_comp} and  \ref{fig:robustness_setup}. 

We chose this proxy task because it requires to distinguish different colors and localize multiple objects, which enables to study DR parameters for color-dependent and multi-object manipulation policies.

To predict 3D locations of the cubes, we use a similar neural network architecture as in Section~\ref{sec:bc} but with RGB frames $\mbf{I}_{t} = \{\mbf{I}^1_{t}, \mbf{I}^2_t \}$ for one time step instead of three and no proprioceptive information at the input.
We train the model by minimizing the distance between the predicted and the ground truth positions of the cubes. 

To evaluate sim-to-real transfer for the proxy task, we train the cube localization network in simulation using different DR parameters and measure localization precision on pre-recorded images of real scenes.
Our empirical experiments in Section~\ref{sec:expr} show that the performance of this proxy task is well aligned with the real-world accuracy of diverse manipulation policies.
\vspace{-.2cm}
\section{Tasks and Datasets}
\vspace{-.1cm}
\label{sec:tasks}

\subsection{Manipulation tasks}
To assess the quality and robustness of our proposed sim-to-real pipeline, we propose a rich set of robotic manipulation tasks with different challenges, see  Figure~\ref{fig:overview}.

\noindent\textbf{Stacking.} The robot has to pick up a red cube on the table and stack it on top of a green cube. Both cubes have a side length of $5 \text{ cm}$. Stacking is most similar to our
proxy task and serves as a starting point to validate our sim-to-real pipeline.

\noindent\textbf{Box-retrieving.} A rectangular gray box of size $20.5\times21.5\times8 \text{ cm}^{3}$ has two compartments containing a green and a red cube respectively. The cubes have a side length of $5 \text{ cm}$. The box is on the table and its lid is closed.
The task requires a gripper to open the box lid, take out the red cube and place it on top of a green marker of size $6 \times 6 \text{ cm}^2$ outside of the box. This task is challenging because it is a long-horizon task that requires a decision which object to pick once the box is opened. 

\noindent\textbf{Assembling.} The task requires grasping a nut with a square hole located at a random position and orientation on the table, and then putting it on a pink rectangular parallelepiped of size $3.5\times3.5\times10.5 \text{  cm}^{3}$ fixed on the table. It requires gripper rotation and high-precision motion to put the nut on the screw.

\noindent\textbf{Pushing.} The goal is to push a green cube of $4 \text{ cm}$ side length with the closed tip of the gripper to reach a square pink marker of $6 \times 6 \text{ cm}^2$ on the table. Pushing requires vision-based reactive behavior and closed-loop control to handle uncertainty of contacts between the gripper, the cube and the table. 

\noindent\textbf{Pushing-to-pick.} A red cube is surrounded by three obstacle cubes with random colors. The goal is to pick the red cube which cannot be performed directly. The gripper needs to first unlock the red cube by pushing two cubes on each side of the red cube. All the cubes are of $5 \text{ cm}$ side length. Pushing-to-pick combines challenges from pushing and picking into a long-horizon task.

\noindent\textbf{Sweeping.} 
The  goal is to sweep $14$ tiny foam cubes of $1.5 \text{ cm}$ side length lying on the table onto a  $16 \times 16 \text{  cm}^2$ green marker. To do so a squared broom of $7 \times 7 \text{ cm}^2$ is attached to the robot gripper.
Sweeping is significantly harder than pushing as it requires to sweep many small objects towards the goal while they can spread. To successfully perform the task, a policy has to perform several sweeps and recover the missing objects that did not reach the target zone, a behavior only obtained with a closed-loop system. 

\noindent\textbf{Rope-shaping.} A deformable rope of length $30  \text{ cm}$ and diameter of $3  \text{ cm}$ is in a random configuration on the table. The gripper needs to manipulate the rope to shape it into a straight line so that the ends of the rope are placed on top of two $6 \times 6 \text{ cm}^2$ pink and green markers. This task is challenging due to the manipulation of a deformable object.

\vspace{-.1cm}
\subsection{Dataset of expert demonstrations}
We leverage object information available in simulation to generate expert trajectories solving the tasks. 
Stacking, pushing-to-pick, box-retrieving and assembling can be solved with an open-loop oracle computing the solution trajectory before execution given the initial gripper and object configuration. 
However, we need to design a closed-loop oracle for pushing, rope-shaping and sweeping tasks, where trajectories need to be computed at each step to deal with the dynamics of the environment. The closed-loop oracles are finite-state machines whose states define a sub-goal to finish the task, e.g., move next to the cube, push the cube, rotate the cube, move to the initial gripper position, etc. At each step, the oracle on a particular state uses the current gripper and object configuration to compute both the next action and next oracle state until the task is completed.
The training datasets consists of $2,000$ demonstrations 
per task for stacking, box-retrieving, pushing and pushing-to-pick, and $4,000$ demonstrations per task for assembling, sweeping and rope-shaping in simulation.
We additionally collect $150$ real robot demonstrations for stacking. 
\vspace{-.15cm}
\section{Experimental Results}
\vspace{-.1cm}
\label{sec:expr}

\begin{figure}[t]
  \centering
  \vspace{0.2cm}\includegraphics[width=\linewidth]{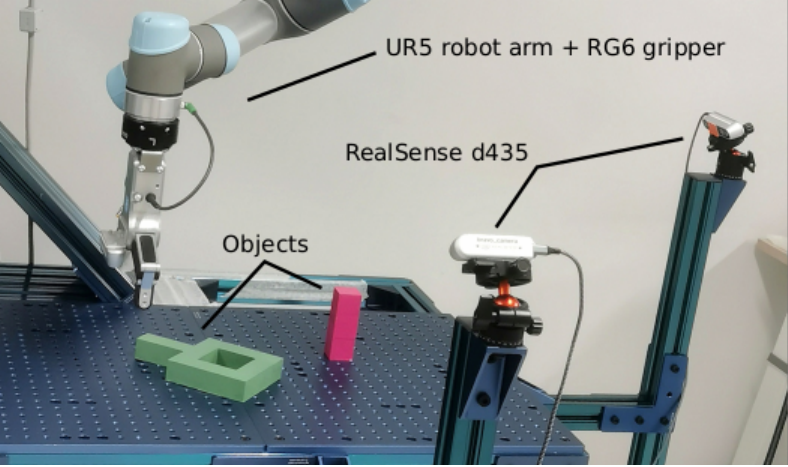}
  \caption{\textbf{Robotic platform.} Our setup includes two RealSense D435 cameras, some objects and the UR5 robotics arm equipped with a RG6 gripper.}
  \vspace{-.6cm}
  \label{fig:experimental_setup}
\end{figure}

\subsection{Experimental Setup}
\label{sec:experimental_setup}

Our robotic platform is a UR5 6-DoF robotic arm equipped with a two-finger Robotiq RG6 gripper, see Figure~\ref{fig:experimental_setup}.
Input RGB images are captured with two Intel RealSense D435 cameras located in the front and the left side of the robotic arm.
We perform control and obtain camera measurements together with proprioceptive information at the frequency of 10 Hz. 
We use the MuJoCo engine \cite{todorov12} to perform physics simulation and rendering of the robotic environment. To align real and simulated images, we carry out extrinsic camera calibration using AprilTag \cite{olson11} and we use the intrinsic parameters provided by the RealSense camera. 
Images are resized and cropped into $180 \times 240$ pixels. 
We define the gripper workspace as a volume of $40\times40\times20 \text{ cm}^{3}$. The gripper initial position is sampled from the set of possible positions inside this workspace while the initial object positions are sampled randomly inside this same workspace and on  the robot table.

\noindent \textbf{Evaluation under real-world variations.}
We evaluate the robustness of our models to natural variations in visual appearance caused by changes in surface textures, lighting and camera poses.  
We consider five scenarios by: (i) adding a textured table cloth, (ii) decreasing the intensity of the room lighting, (iii) adding a variable color lighting with two Olafus $50 \text{W}$ RGB LED flood lights, (iv) offseting object colors and (v) adding noise to the camera pose. Figure~\ref{fig:robustness_setup} illustrates the default scenario as well as the five variations.

\subsection{Implementation details}

\begin{figure}[t]
\vspace{0.2cm}
  \centering
  \includegraphics[width=\linewidth]{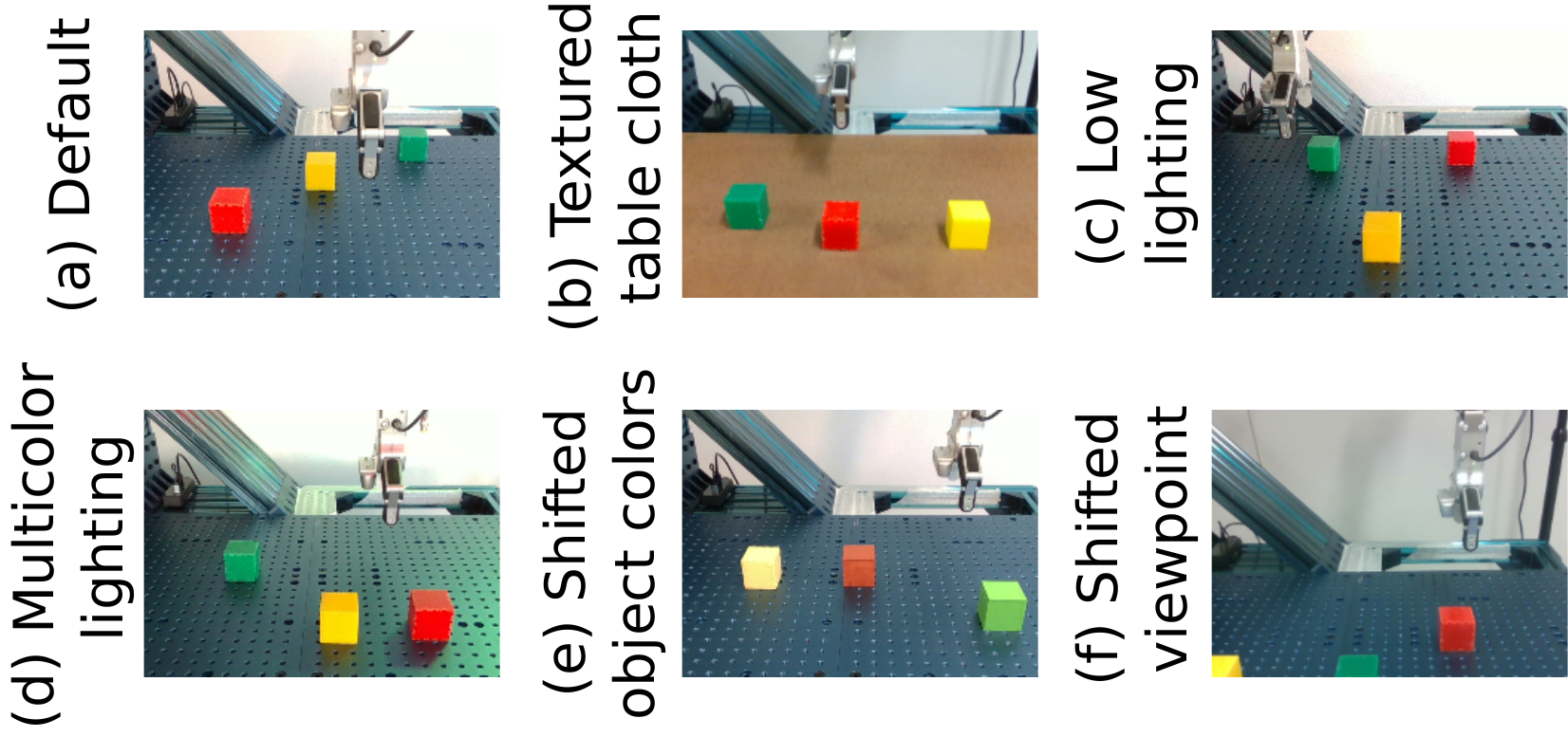}
  \caption{\textbf{Scenarios with varying visual appearance.} 
  We study the robustness of our models to variations in visual appearance caused by changes in surface texture (b), lighting~(c, d), object colors (e) and cameras pose (f).}
  \vspace{-.7cm}
  \label{fig:robustness_setup}
\end{figure}

\noindent\textbf{Cube localization proxy task.} To evaluate the precision of cube localization, we use three cubes with side length of $5\text{cm}$ and colored green, red and yellow respectively. We collect a real dataset for the proxy task where we sample random positions for the cubes and the gripper inside the environment workspace. 
We repeat this 250 times and vary the viewing conditions, such as background texture, lightening and camera viewpoint, see Figure~\ref{fig:robustness_setup}. 
While in simulation we know the cubes and gripper position, on the real robot we do not have direct access to this information. 
To overcome this limitation, we manually place the cubes in known positions, and the gripper moves each cube to a random position. Finally, the gripper moves to a random position, and we use the robot forward kinematics at each step to create the ground truth. 
We train our model for $400,000$ gradient steps using the AdamW optimizer \cite{loshchilov18}. We use a cosine scheduler with initial learning rate of $3 \times 10^{-4}$ and the minimum learning rate of $1 \times 10^{-6}$.

\noindent\textbf{Policy learning.} 
We optimize robotic manipulation policies using a batch size of $32$. 
For the stacking, pushing, pushing-to-pick and box-retrieving tasks we perform $400,000$ gradient steps while the more difficult tasks of assembling, rope-shaping and sweeping are optimized for $1\text{M}$ steps. 
We evaluate the performance of our models on $250$ episodes in simulation with domain randomization disabled, and run $20$ trials for each model and task when performing real robot evaluation. 
We experimentally set the value of $\lambda$ in (\ref{eqn:loss}) to $0.8$ during training and train our policy models with the AdamW optimizer~\cite{loshchilov18} and a cosine scheduler.

\begin{table*}[t]
\small
\vspace{0.2cm}
\setlength\tabcolsep{2.7pt}
\centering
\begin{tabular}{lcccccccc}
  \toprule
  Model & Stacking & Box-retrieving & Assembling & Pushing & Pushing-to-pick & Sweeping & Rope-shaping & Average \\
  \midrule
  Baseline & 39.6 & 67.2 & 14.0 & 96.0 & 26.8 & 28.4 & 38.4 & 44.97 \\
  + Second viewpoint & 61.6 & 83.6 & 64.4 & 95.2 & 64.0 & 17.2 & 72.0 & 65.43 \\
  + Multiple frames & 96.4 & 100.0 & 98.8 & 94.8 & 98.8 & 80.0 & 88.8 & 93.94 \\
  + Proprioceptive info. & 100.0 & 100.0 & 99.6 & 95.6 & 100.0 & 98.0 & 95.2 & 98.34 \\
  \bottomrule
\end{tabular}
\caption{\textbf{Comparison of different modeling choices for policy learning in simulation.} We train on 2000 demonstrations for stacking, box-retrieving, pushing and pushing-to-pick and 4000 for assembling, sweeping and rope-shaping. We evaluate models on 250 episodes for each manipulation task and report the success rate (\%).}
  \vspace{-.2cm}
\label{table:policy_ablation}
\end{table*}

\begin{figure*}[t]
\begin{subfigure}{.24\textwidth}
    \centering
  \includegraphics[width=\linewidth]{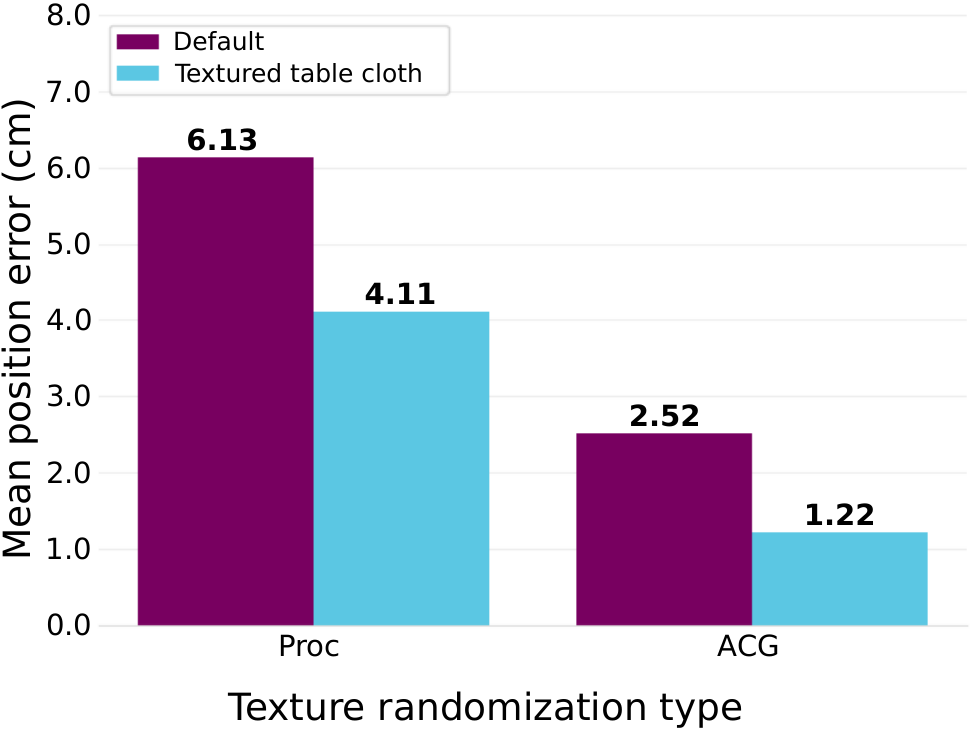}
  \caption{\textbf{Texture types. }}
  \label{fig:texture_study}
\end{subfigure}
\begin{subfigure}{.24\textwidth}
  \centering
  \includegraphics[width=\linewidth]{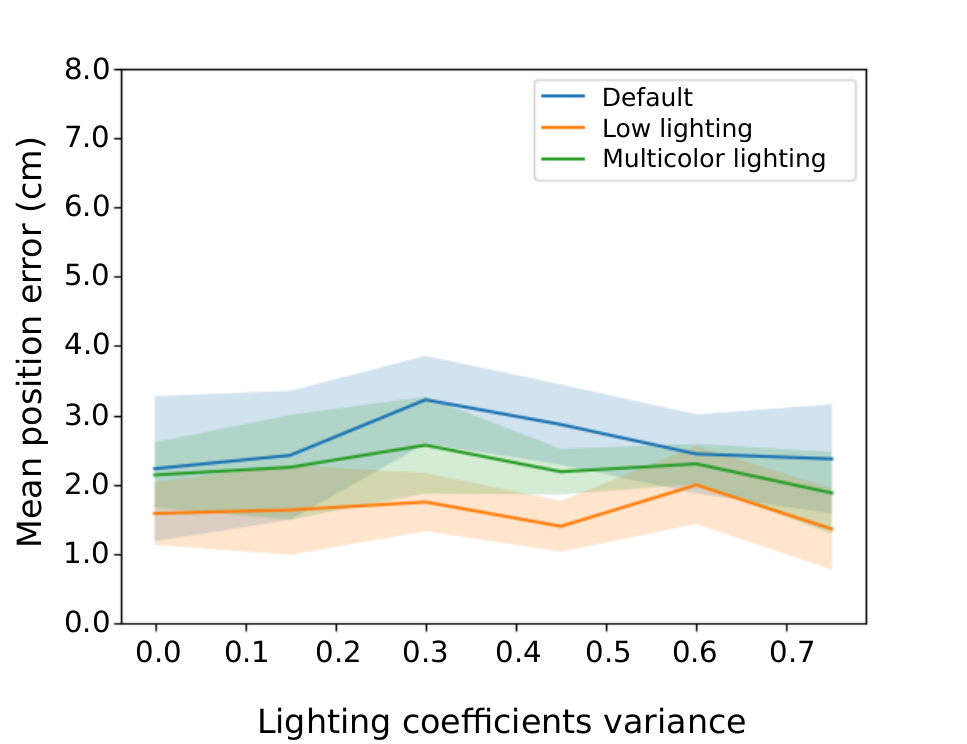}
  \caption{\textbf{Light parameters $\phi_l$.}}
  \label{fig:light_study}
\end{subfigure}
\begin{subfigure}{.24\textwidth}
    \centering
  \includegraphics[width=\linewidth]{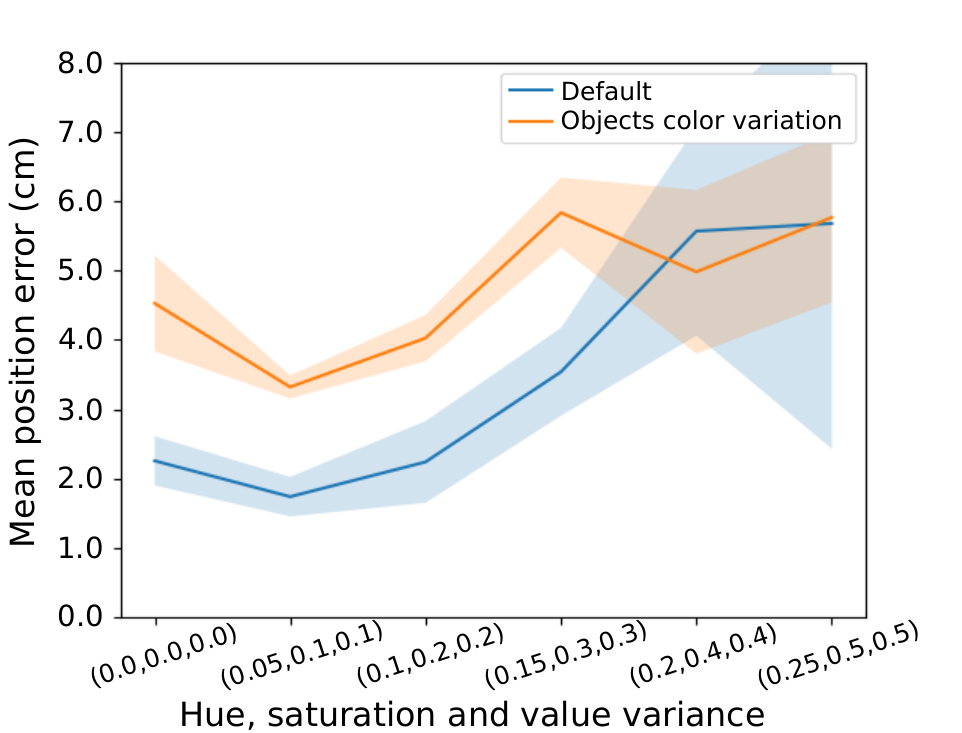}
  \caption{\textbf{Object color parameters $\phi_o$.}}
  \label{fig:color_study}
\end{subfigure}
\begin{subfigure}{.24\textwidth}
    \centering
  \includegraphics[width=\linewidth]{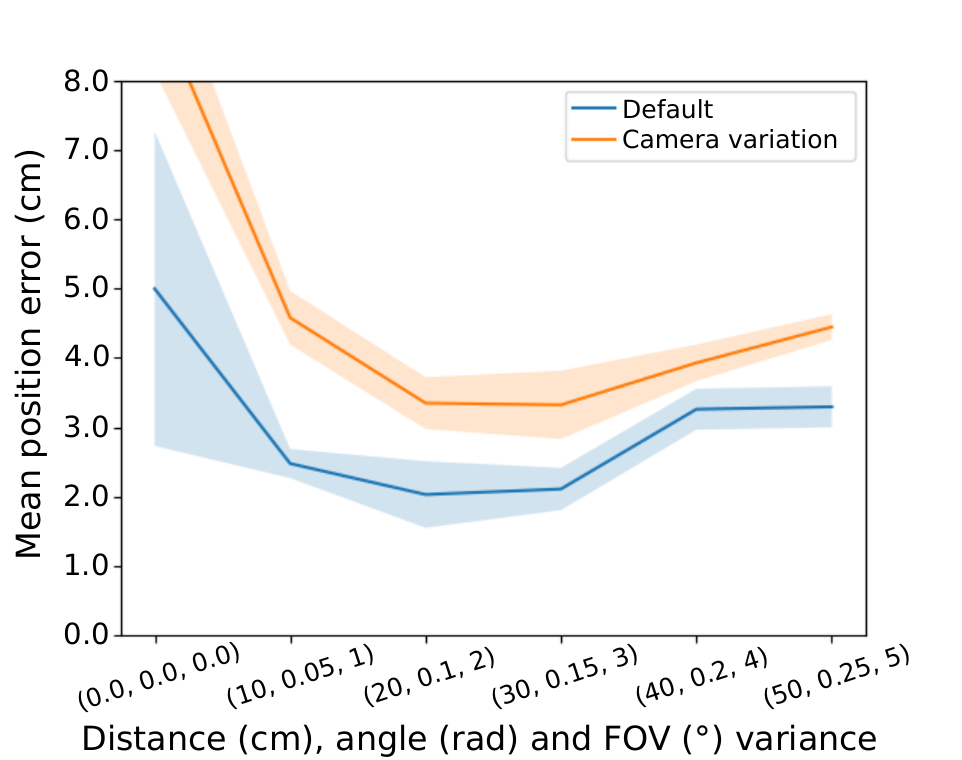}
  \caption{\textbf{Camera parameters.}}
  \label{fig:cam_study}
\end{subfigure}
\caption{\textbf{Impact of domain randomization parameters on the proxy cube localization task.} The performance (y-axis) is mean position error (cm) over 250 images in real scenes, where the lower the better.}
\label{fig:dr_params}
\vspace{-.5cm}
\end{figure*}

\subsection{Policy learning in simulation}
\label{sec:policy_sim}

In this section, we evaluate the design choices of our policy network in simulation for all seven tasks, see 
Table~\ref{table:policy_ablation}. We train our models on 2000 (4000) demonstrations in simulation using DR with the parameters defined in Section~\ref{sec:dr_ablation} and evaluate them in simulation without any augmentation.
The baseline takes as input a single viewpoint and a single frame and reaches 44.97\% success rate on average.
Adding a second viewpoint improves the performance to $65.43\%$ on average and results in $50.4\%$ absolute gain for the assembling task which requires high precision. 
Adding a short-term history by taking as input three frames results in a further improvement of $28.51\%$ on average. 
Finally, we add proprioceptive information to the input, which further improves the performance by $4.4\%$ and leads to the best performance of $98.34\%$ success rate.
Propioceptive information is essential for tasks such as sweeping (+$18\%$) where a high precision of the gripper position is required. 

\subsection{Domain randomization ablation with a proxy task}
\label{sec:dr_ablation}

In this section, we first determine the best domain randomization (DR) parameters for each augmentation type on the proxy localization task.
Figure~\ref{fig:texture_study} compares the two different texture types and shows that ambientCG textures are much more effective than the procedurally generated ones. As illustrated in Figure~\ref{fig:textures_comp}, ambientGC textures are of high quality and have richer variations than procedurally generated textures which could explain this difference in error.
In the following, we always use ambientCG texture randomization.
For lighting randomization, Figure~\ref{fig:light_study} shows that performance is not very sensitive to different randomization ranges for light ambiance, diffuse and specular coefficients. Hence, we select an intermediate value randomizing the previous three coefficients in the range $[0, 0.6]$. 
For the object colors, HSV color offset has a significant impact as shown in Figure~\ref{fig:color_study}. 
A large offset significantly deteriorates the performance as it confuses colors of different objects. If the offset is too small, the model is not robust enough. We use an offset around the object nominal color of $\phi_o=(0.05, 0.1, 0.1)$ that achieves the best performance.
Finally, as shown in Figure~\ref{fig:cam_study} a small randomization of camera parameters is important due to possible calibration errors and too much variation hurts the performance. We offset the values for camera position by $[-10, 10]\text{ cm}$, for camera angle by $[-0.05, 0.05]\text{ rad}$ and for field of view by $[-1.0, 1.0]^{\circ}$.

\begin{table}
\centering
\tabcolsep=0.1cm
\begin{tabular}{ccccccccc} \toprule
\multicolumn{2}{c}{Training data} & \multirow{2}{*}{\begin{tabular}[c]{@{}c@{}}Img\\ aug\end{tabular}} & \multirow{2}{*}{\begin{tabular}[c]{@{}c@{}}Texture\\ ACG\end{tabular}} & \multirow{2}{*}{Light} & \multirow{2}{*}{\begin{tabular}[c]{@{}c@{}}Obj\\ color\end{tabular}} & \multirow{2}{*}{Camera} & \multicolumn{2}{c}{Error (cm)} \\
Synt & Real &  &  &  &  &  & default & variations \\ \midrule
\multirow{7}{*}{20k} & \multirow{7}{*}{-} & $\times$ &  $\times$ & $\times$ & $\times$ & $\times$ & 7.55 & 8.22 \\
  & & \checkmark & $\times$ & $\times$ & $\times$ & $\times$ & 6.92 & 7.17 \\
  &  & $\times$ & \checkmark & $\times$ & $\times$ & $\times$ & 2.52 & 3.53  \\
  &  & $\times$ & \checkmark & \checkmark & $\times$ & $\times$ & 2.66 & 3.65 \\
  &  & $\times$ & \checkmark & \checkmark & \checkmark & $\times$ & 1.62 & 2.92 \\
  &  & $\times$ & \checkmark & \checkmark & \checkmark & \checkmark & 1.33 & 2.70 \\
 &  & \checkmark & \checkmark & \checkmark & \checkmark & \checkmark & 0.95 & 1.97 \\
100k & - & \checkmark & \checkmark & \checkmark & \checkmark & \checkmark & {\bf 0.48} & {\bf 1.39}  \\ \midrule 
- & 750 & $\times$ & $\times$ & $\times$ & $\times$ & $\times$ & 0.72 & 3.08 \\
100k & 750 & \checkmark & \checkmark & \checkmark & \checkmark & \checkmark & {\bf 0.14} & {\bf 1.00} \\ \bottomrule
\end{tabular}
\caption{\textbf{Domain randomization ablation on the cube localization proxy task.}
The training data include 750 real images, 20k and 100k synthetic images, as well as synthetic images followed by real images fine-tuning.
The models are evaluated on 250 real images without (default) and with scene appearance variation (variations).
We report the localization error (cm).}
\label{table:dr_ablation}
  \vspace{-.8cm}
\end{table}

\begin{table*}[t]
\vspace{.3cm}
\centering
\tabcolsep=0.13cm
\begin{tabular}{ccccccccccccc} \toprule
  \multirow{2}{*}{\begin{tabular}[c]{@{}l@{}}img\\ aug\end{tabular}} & \multicolumn{1}{c}{Texture} & \multirow{2}{*}{Light} & \multirow{2}{*}{\begin{tabular}[c]{@{}c@{}}Obj\\ color\end{tabular}} & \multirow{2}{*}{Camera} & \multicolumn{8}{c}{Success rate} \\
  & ACG &  &  &  & Stacking & Box-retrieving & Assembling & Pushing  & Pushing-to-pick & Sweeping & Rope-shaping & Average \\ \midrule 
 \checkmark & $\times$ & $\times$ & $\times$ & $\times$ & 0/20 & 0/20 & 0/20 & 0/20 & 0/20 & 0/20 & 0/20 & 0/20 \\
 \checkmark & \checkmark & $\times$ & $\times$ & $\times$ & 15/20 & 16/20 & 7/20 & 13/20 & 11/20 & 9/20 & 8/20 & 11.3/20 \\ 
 \checkmark & \checkmark & \checkmark & \checkmark & $\times$ & 17/20 & 15/20 & 13/20 & 18/20 & 13/20 & 11/20 & 11/20 & 14.0/20 \\ 
  \checkmark & \checkmark & \checkmark & \checkmark & \checkmark & \textbf{20/20} & \textbf{17/20} & \textbf{16/20} & \textbf{18/20} & \textbf{17/20} & \textbf{19/20} & \textbf{18/20} & \textbf{18.6/20} \\ 
\bottomrule
\end{tabular}
\caption{\textbf{Comparison of different domain randomizations.} We report performance on seven manipulation tasks with a policy trained in simulation and evaluated on a real robot. For each task we run 20 episodes on the robot and report the success rate.}
\label{table:policy_success}  
\vspace{-.6cm}
\end{table*}

Table \ref{table:dr_ablation} compares combinations of different DR types using the selected parameters. We evaluate both the error on the proxy localization task for the default scene appearance and the average error of all the varied appearance scenes, see Figure~\ref{fig:robustness_setup}.
In addition, we compare DR with the standard 2D image augmentation; we follow \cite{lee21} and modify synthetic images using a random offset $[-0.12, 0.12]$ for brightness, $[-0.05, 0.05]$ for hue, $[0.5, 1.5]$ for saturation, $[0.5, 1.5]$ for contrast and $[-4, 4]$ pixels for horizontal and vertical image translation. 

The first row shows the results for training with 20k synthetic images. We can observe that by adding image augmentations the performance improves only very slightly. 
Texture randomization significantly improves the performance and decrease the error from $7.55$ to $2.52 \text{ cm}$.
Adding lighting randomization does not improve performance, i.e., the error remains similar. 
With the variation of object colors, we further improve the error by $1\text{ cm}$ and achieve an error of $1.62 \text{ cm}$.
Varying the camera parameters additionally improves performance to $1.33 \text{ cm}$. 
Further adding 2D image augmentations has a minor impact on the error and decreases it to $0.95 \text{ cm}$. 

Since we are using synthetic data for training, we can easily generate additional data.  We show a large boost in performance when increasing the dataset from $20,000$ to $100,000$ images, resulting in an error of $0.48 \text{ cm}$. Notably, our model trained with DR and simulation-only data outperforms the model trained on real but limited data, and is also more robust to variations in the scene appearance. The error for such data with variations (last column of Table~\ref{table:dr_ablation} is $1.39 \text{ cm}$ 
 for synthetic training data with DR in contrast to $3.08 \text{ cm}$ obtained when training with real data.
Finally, we show that additional fine-tuning of our model on real data is beneficial, i.e., the error decreases to  $0.14 \text{ cm}$.

\subsection{Ablation of domain randomization for policy learning}
\label{sec:policy_real}

In this section, we measure the impact of domain randomization (DR) on policy learning beyond the proxy task.
We train policies with our DR approach on a suite of seven manipulation tasks described in Section~\ref{sec:tasks}.
The same DR parameters selected for the cube localization proxy task are used for all the manipulation tasks.

Table~\ref{table:policy_success} presents the results. We compare our full DR approach with three variations of Table~\ref{table:dr_ablation}. The first variation uses only 2D image augmentation; the second one adds ambientCG texture randomization to the 2D image augmentation; and the third one includes lighting and object color randomization on top of the previous variation. 
Using only 2D image augmentation is not enough to succeed in any task on the real robot, i.e., the success rate is 0. 
Adding ACG texture randomization helps transferring the policy with an average success rate of $11.3/20$. Additionally adding light and object color randomization improves the performance further and results in an average success rate of $14.0/20$. 
Adding small camera variations to the training further improves the performance, i.e.,  our full DR setup obtains a success rate of $18.6/20$ on average. 
When comparing results in Tables~\ref{table:dr_ablation} and~\ref{table:policy_success}, we can see that \emph{the performance of the proxy task is well aligned with the performance of policies for all considered manipulation tasks on the real robot}. 
This validates the effectiveness of our proxy task for choosing DR parameters for the visual sim-to-real policy transfer. We illustrate successful runs of our policies on the real robot in Figure~\ref{fig:tasks_real}. 
Additional qualitative results for all seven manipulation tasks are presented in the supplementary video. 

\begin{figure}[t]
  \centering
  \includegraphics[width=\linewidth]{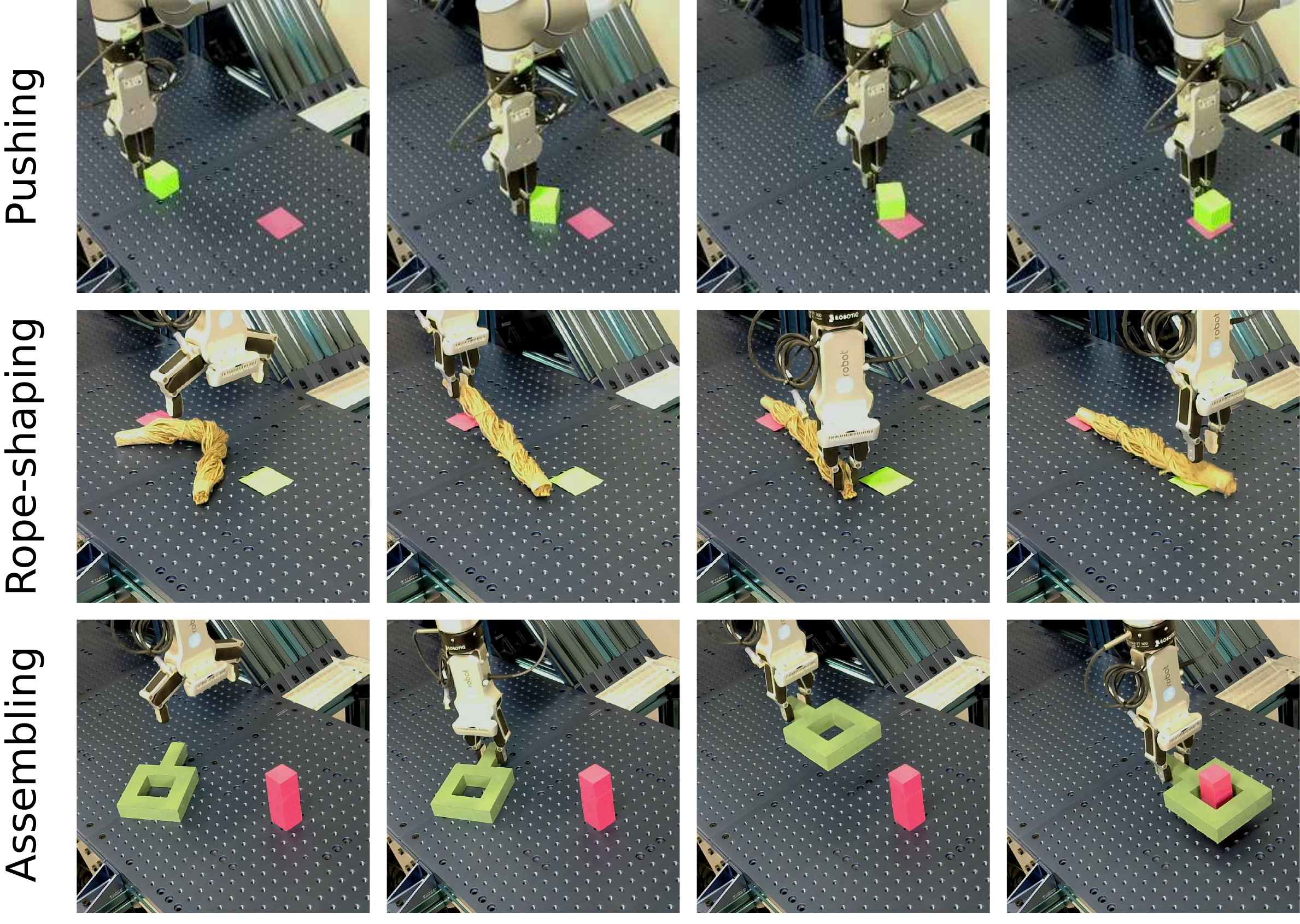}
  \caption{\textbf{Real robot experiments.} We illustrate successful runs for pushing, rope-shaping and assembling.}
  \label{fig:tasks_real}
  \vspace{-0.3cm}
\end{figure}
\begin{figure}[t]
  \centering
  \includegraphics[width=\linewidth]{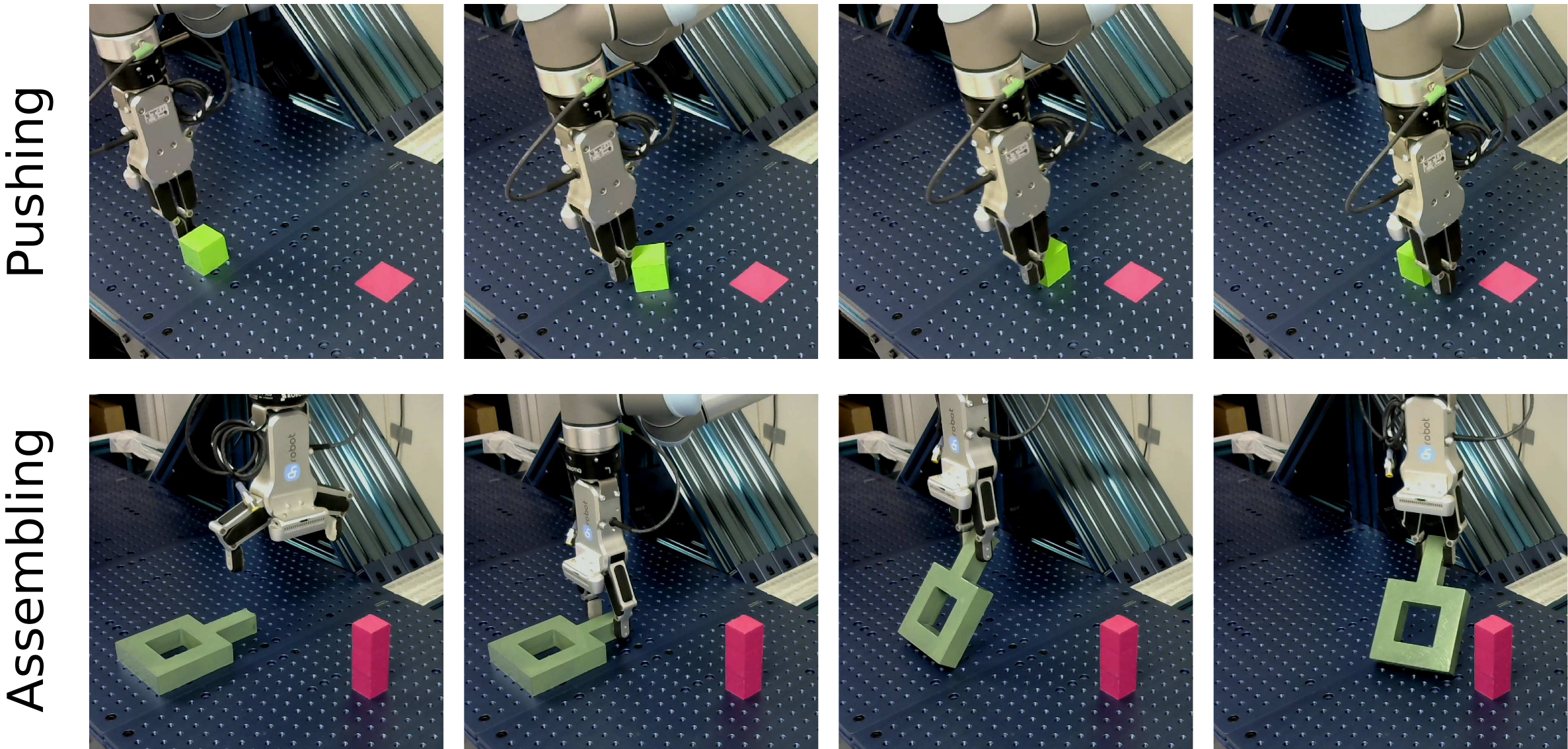}
  \caption{\textbf{Real robot experiments.} We illustrate failure cases for pushing and assembling.}
  \label{fig:tasks_real_failure}
  \vspace{-0.7cm}
\end{figure}

\noindent \textbf{Failure cases analysis.}
 Figure~\ref{fig:tasks_real_failure} illustrates two type of failure cases. The first type of failure (top row) is due to unseen states. During pushing the cube rotates by 90 degrees, a state never observed in training. This causes the gripper to leave the pushing surface and continue the pushing to the target place without the cube.
The second one originates from the physical sim-to-real gap (bottom row). During assembling the gripper grasps the nut by the end of the handle and the nut rotates due to its weight and the lack of sufficient friction between the gripper fingers and the nut. This issue does not occur in simulation due to the insufficient realism in friction modeling.

\subsection{Robustness of policies to visual variations}

In this section, we evaluate the robustness to visual variations for the stacking task on the real robot. Table~\ref{table:robustness_policy} compares  our policy trained with DR, a policy trained on real data and a policy trained on DR and fine-tuned on real data for the different variations described in Section~\ref{sec:experimental_setup}. 
We use $2,000$ demonstrations in simulation and $150$ real robot demonstrations. While simulation allows us to gather $2000$ demonstrations in minutes, gathering $150$ demonstrations on the real robot is time-consuming (approx.~4.5 hours).

\begin{table}[t]
\vspace{0.2cm}
\small
\centering
\begin{tabular}{lrrr}
  \toprule
   & DR & Real & DR+Real \\
  \midrule
  Default & 20/20 & 20/20  & 20/20 \\
  Textured table cloth & 20/20 & 1/20 & 20/20 \\
  Low lighting & 19/20 & 17/20 & 20/20 \\
  Multicolor lighting & 16/20 & 14/20 & 17/20 \\
  Object colors variation & 18/20 & 19/20 & 20/20 \\
  Camera variation & 11/20 & 8/20 & 12/20 \\
  \midrule
  Average & 17.3/20 & 13.2/20 & \textbf{18.2/20} \\
  \bottomrule
\end{tabular}
\caption{\textbf{Success rate for stacking on a real robot under default and five variations of the scene appearance.} We compare policies trained with synthetic data with domain randomization (2000 demonstrations + DR), real data (150 real demonstrations), and their combination (2000 demonstration + DR and fine-tuned on 150 real demonstrations). 
}
\vspace{-.6cm}
\label{table:robustness_policy}
\end{table}

In the default setting without visual appearance variation of real scenes, all the policies achieve $100\%$ success rate. However, our DR policy exhibits much stronger robustness to changes in the table texture and achieves $100\%$ success rate compared to $5.0\%$ success rate obtained by the real data policy. 
Similarly, variations of the lighting conditions caused by lowering light intensity and changing light color result in larger decrease of performance for policies trained on the real data compared to our policies. 
When we modify cube colors, we can see that both the DR and real data policy perform well achieving a success rate of $90.0 \%$ and $95.0 \%$ respectively.
Finally, the variation of camera parameters results in a large degradation of both policies.
Overall we conclude that policies trained on synthetic data with DR are more robust compared to policies trained on the real data, when such data is available in limited amounts. Additionally, using the real data to fine-tune the policies trained with DR results in an additional increase of the average success rate of $+4.5 \%$.

\vspace{-.1cm}

\section{Conclusion}
\vspace{-.1cm}
In this paper we address visual sim-to-real transfer of manipulation policies using domain randomization.
We explore multiple visual domain randomization components and propose a proxy task of cube localization to choose appropriate DR parameters. 
We demonstrate that under the same DR settings, the real-world performance on our proxy task strongly correlates with the performance of policies trained for diverse and challenging manipulation tasks. 
We introduce a rich set of seven manipulation tasks to benchmark visual sim-to-real transfer and demonstrate that our method significantly and consistently outperforms other approaches without DR, achieving 93\% average success rate on a real robot.
Our method also demonstrates increased robustness to visual appearance changes in real scenes.

\vspace{.1cm}

{
\small
\textbf{Acknowledgements.}
This work was partially supported by
the HPC resources from GENCI-IDRIS (Grant 20XX-AD011012122). 
It was funded in part by the French government under management of Agence Nationale de la Recherche as part of the “Investissements d’avenir” program, reference ANR19-P3IA-0001 (PRAIRIE 3IA Institute), the ANR project VideoPredict (ANR-21-FAI1-0002-01) and by Louis Vuitton ENS Chair on Artificial Intelligence.
}

\vspace{-.2cm}
\bibliographystyle{IEEEtran}
\bibliography{bibliography}
\end{document}